\documentclass[10pt,twocolumn,letterpaper]{article}

\usepackage{cvpr}              

\definecolor{cvprblue}{rgb}{0.21,0.49,0.74}
\usepackage[pagebackref,breaklinks,colorlinks,allcolors=cvprblue]{hyperref}
\usepackage{subcaption, stackengine}
\usepackage{algpseudocode}
\usepackage{orcidlink}
\usepackage{algorithm}
\usepackage{bbm}

\def\paperID{*****} 
\def\confName{CVPR}
\def\confYear{2025}

\title{Deep Diffusion Models and Unsupervised Hyperspectral Unmixing for Realistic Abundance Map Synthesis}

\author{Martina Pastorino\orcidlink{0000-0002-3804-4768}\\
DITEN, University of Genoa\\
16145 Genoa, Italy\\
{\tt\small martina.pastorino@unige.it}
\and
Michael Alibani\orcidlink{0000-0003-0023-4969}\\
DII, University of Pisa\\
56122 Pisa, Italy\\
{\tt\small michael.alibani@unipi.it}
\and 
Nicola Acito\orcidlink{0000-0003-1984-7992}\\
DII, University of Pisa\\
56122 Pisa, Italy\\
{\tt\small nicola.acito@unipi.it}
\and
Gabriele Moser\orcidlink{0000-0002-3796-2938}\\
DITEN, University of Genoa\\
16145 Genoa, Italy\\
{\tt\small gabriele.moser@unige.it}
}

\begin{document}
\maketitle
\begin{abstract}
This paper presents a novel methodology for generating realistic abundance maps from hyperspectral imagery using an unsupervised, deep-learning-driven approach. Our framework integrates blind linear hyperspectral unmixing with state-of-the-art diffusion models to enhance the realism and diversity of synthetic abundance maps. First, we apply blind unmixing to extract endmembers and abundance maps directly from raw hyperspectral data. These abundance maps then serve as inputs to a diffusion model, which acts as a generative engine to synthesize highly realistic spatial distributions. Diffusion models have recently revolutionized image synthesis by offering superior performance, flexibility, and stability, making them well-suited for high-dimensional spectral data. By leveraging this combination of physically interpretable unmixing and deep generative modeling, our approach enables the simulation of hyperspectral sensor outputs under diverse imaging conditions—critical for data augmentation, algorithm benchmarking, and model evaluation in hyperspectral analysis. Notably, our method is entirely unsupervised, ensuring adaptability to different datasets without the need for labeled training data. We validate our approach using real hyperspectral imagery from the PRISMA space mission for Earth observation, demonstrating its effectiveness in producing realistic synthetic abundance maps that capture the spatial and spectral characteristics of natural scenes.
\end{abstract}    
\section{Introduction}
\label{sec:intro}

Hyperspectral imaging has emerged as a powerful tool in various scientific and industrial applications, because of its very high spectral resolution. By capturing detailed spectral information across a broad range of wavelengths, hyperspectral sensors provide valuable insights into the composition and characteristics of materials present in a scene. This capability has enabled advances in multiple domains, including remote sensing~\cite{richards}, environmental monitoring~\cite{Zhang18032024}, urban planning~\cite{9773336}, precision agriculture~\cite{10353994}, and biomedical imaging for disease diagnosis and image-guided surgery~\cite{PMC3895860}. However, despite its potential, the widespread adoption of hyperspectral imaging is hindered by some challenges, as their acquisition may be constrained by environmental factors such as atmospheric conditions and sensor limitations. Additionally, real-world hyperspectral datasets may suffer from noise, radiometric distortions, and spatial inconsistencies, further complicating their use for algorithm development and validation.

To mitigate these limitations, synthetic data generation has become an increasingly relevant approach. The ability to produce realistic hyperspectral datasets allows researchers to simulate sensor performance, support the planning of space missions, and generate large-scale training data for machine learning models. A crucial aspect of hyperspectral image simulation is the generation of abundance maps, which describe the fractional composition of materials within each pixel. Since hyperspectral images result from the interaction of multiple materials within a given spatial resolution, abundance maps provide an interpretable representation of the scene content. Instead of synthesizing hyperspectral data directly in the high-dimensional spectral space, the simulation task can be reformulated in the abundance space, reducing computational complexity while preserving the statistical and spatial characteristics of real data.

Recent advances in generative modeling have provided robust frameworks for synthetic data generation, particularly with the rise of foundation generative models~\cite{bommasani_arxiv}. These large-scale models, trained on vast and diverse datasets, have demonstrated impressive generalization capabilities in various domains, including text~\cite{brown_nips}, images~\cite{Radford2021LearningTV, dm, dm_icml}, and even hyperspectral data synthesis~\cite{unmixingDM}. Foundation generative models, such as diffusion models and transformer-based architectures, offer scalable and versatile solutions for data generation by capturing complex distributions with minimal supervision. In the context of hyperspectral imaging, such models can be adapted to learn meaningful spectral-spatial representations~\cite{spectral_gpt}, facilitating the creation of realistic synthetic datasets. By leveraging the strengths of these generative paradigms, our approach integrates domain-specific constraints with state-of-the-art generative modeling to enhance the realism and usability of synthetic abundance maps.

A key approach for abundance estimation is blind linear hyperspectral unmixing, which aims to decompose hyperspectral images into their constituent spectral signatures, known as endmembers, and their corresponding abundance maps. Unlike supervised unmixing methods, which rely on prior knowledge of endmember spectra, blind unmixing methods estimate both the spectral signatures and abundances directly from the data~\cite{hysupp}. Given the importance of unmixing in hyperspectral analysis, numerous techniques have been developed, utilizing statistical~\cite{MV_bum}, Bayesian~\cite{BSMA}, and deep learning-based methodologies~\cite{cnnaeu, pgmsu}. These approaches seek to improve the accuracy and robustness of abundance estimation, which is a crucial step in hyperspectral image interpretation and synthesis.

In recent years, diffusion models have emerged as state-of-the-art generative models for image synthesis~\cite{bishop, dm, unmixingDM, dm_gan, dm_icml}. Unlike traditional generative adversarial networks (GANs)~\cite{gan, cyclegan} or autoencoders~\cite{luppino}, which directly learn mappings from latent representations to images, diffusion models operate by progressively transforming random noise into structured images through a stepwise denoising process. This approach, inspired by stochastic differential equations, has demonstrated remarkable success in generating high-quality samples while avoiding common pitfalls of GANs, such as mode collapse and training instability~\cite{bishop, dm_gan}. By learning to reverse a noise corruption process, diffusion models effectively capture complex spatial structures and realistic feature distributions, making them well-suited for hyperspectral data synthesis.

In this work, we propose an unsupervised methodology for generating synthetic abundance maps derived from real hyperspectral imagery. The approach integrates blind linear unmixing for extracting spectral endmembers and abundance maps with a diffusion model for generating realistic spatial distributions that mimic the patterns observed in real-world hyperspectral data. The unmixing process provides a physically interpretable decomposition of hyperspectral signals, while the diffusion model ensures that the generated abundance maps exhibit realistic spatial coherence and material distributions. By leveraging hyperspectral datasets from the PRISMA mission, we ensure that the synthetic abundance maps maintain spectral fidelity and physical plausibility, making them useful for hyperspectral image simulation, domain adaptation, sensor performance assessment, and data augmentation for machine learning applications. 

Compared to the previous method in~\cite{unmixingDM}, we do not propose a specific autoencoder-based unmixing method to produce the abundance maps used to train the diffusion model. We rather use a set of existing hyperspectral unmixing methods, drawing from the three methodological families (statistical, deep learning, and least squares-based), in order to obtain heterogeneous abundance maps. This allows for more generality: the generated output by the diffusion model is not conditioned to a single unmixing algorithm and its possible specific limitations, but is influenced by the dictionary of methods.

This paper is organized as follows. Section~\ref{sec:meth} describes the proposed methodology, detailing the integration of blind unmixing techniques and diffusion-based generative modeling. Section~\ref{sec:exp} presents the experimental validation, showcasing synthetic abundance maps generated for different land cover types, including mountainous, vegetated, urban, and mixed environments. The effectiveness of the proposed approach is evaluated through qualitative visual analysis, demonstrating its potential for advancing hyperspectral image simulation and providing reliable synthetic datasets for remote sensing research and beyond.

\section{Proposed Method}\label{sec:meth}
\subsection{Overview of the proposed approach}
\begin{figure*}
    \centering
    \includegraphics[width=\linewidth]{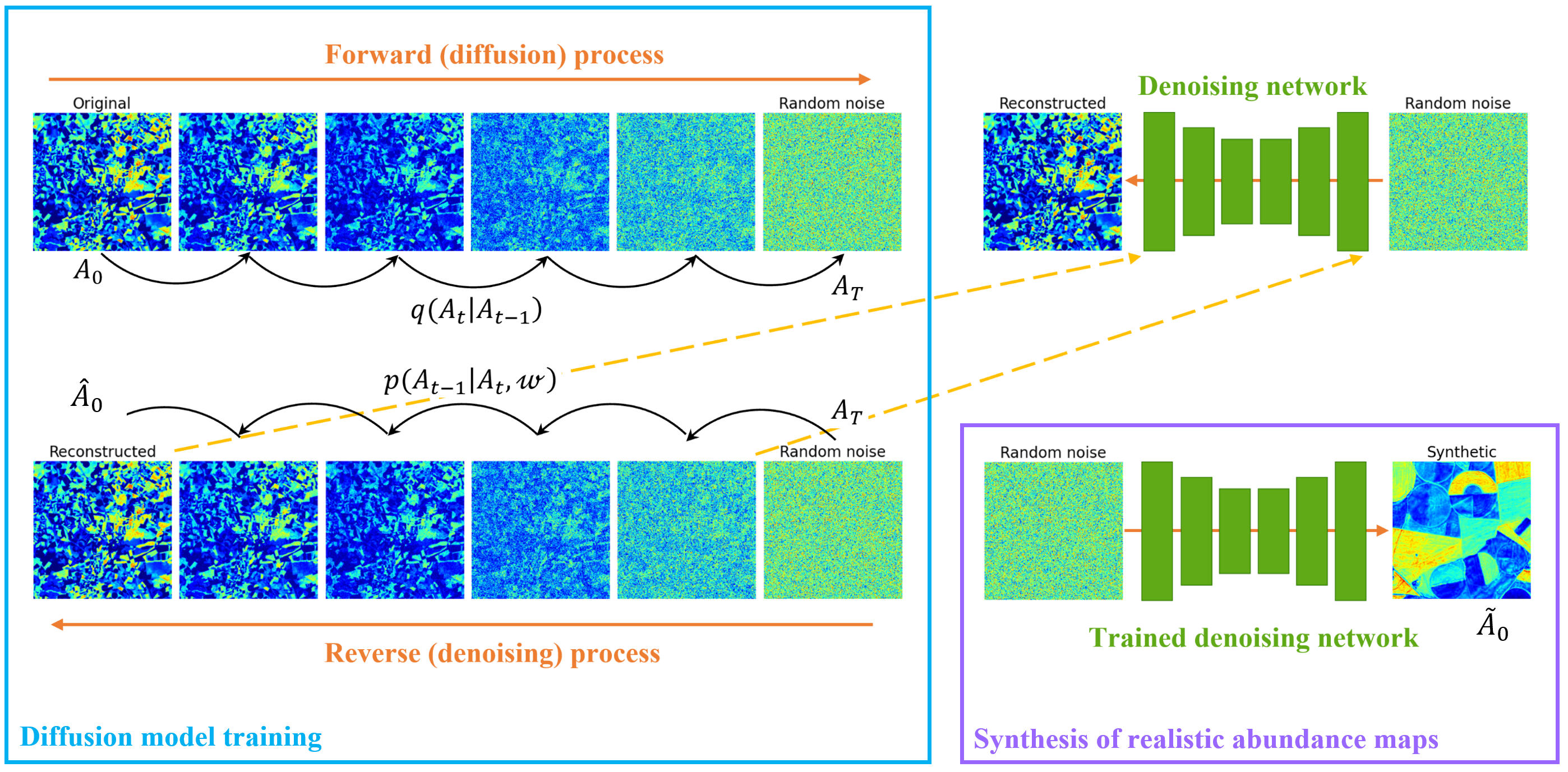}
    \caption{Architecture of the proposed method for the synthesis of realistic abundance maps.}
    \label{fig:arch}
\end{figure*}

This work is framed within the broader scope of hyperspectral data simulation, a challenging task due to the high spectral dimensionality of hyperspectral imagery. Directly synthesizing high-fidelity hyperspectral data remains difficult; therefore, we propose an intermediate synthesis approach that focuses on generating realistic abundance maps. By addressing the synthesis problem in the abundance space—where the dimensionality is significantly reduced—we mitigate the complexity of direct hyperspectral generation.

Consequently, our method incorporates an initial hyperspectral unmixing step. The goal is to estimate the endmembers and abundance maps represented by the hyperspectral acquisition. Hyperspectral unmixing algorithms aim to extract the pure spectral signature of the materials (endmembers) constituting the surface captured in an image, and the corresponding fractional abundances of each of these signatures per pixel (abundances). Afterwards, the estimated abundance maps are used to train a diffusion model, whose final goal is to synthetize realistic abundance maps which could be captured by a hyperspectral sensor. The overall methodology is shown in Fig.~\ref{fig:arch}.

\subsection{Blind hyperspectral unmixing}

In particular, we focus on linear unmixing methods within an unsupervised -- or "blind" -- setup, where the endmember signatures are not known a priori and must be estimated jointly with the corresponding abundance maps. Blind unmixing is crucial in scenarios where spectral libraries are unavailable or unreliable, such as planetary exploration, mineral mapping, and remote sensing of novel or evolving environments. Compared to supervised methods, which require predefined spectral signatures, blind unmixing is more flexible but also significantly more challenging due to the ill-posed nature of the problem and its sensitivity to initialization~\cite{hysupp}. In the linear case, the problem is formulated using a linear mixture model as follows:

\begin{equation}
    Y = E \cdot A + N \quad s.t. \quad A\geq 0, A^T \cdot \mathbbm{1}_p = \mathbbm{1}_S 
\end{equation}
where $Y\in \mathbb{R}^{C \times S}$ is the original hyperspectral image, with $S$ pixels and $C$ spectral channels. The matrix $E \in \mathbb{R}^{C\times p}$ contains the spectral signatures of $p$ endmembers for each $C-$dimensional pixel in $Y$, while $A \in \mathbb{R}^{p\times S}$ is the matrix of the fractional abundances for each endmember $p$, and $N$ is an additive noise term. The matrix inequality $A\geq 0$ expresses elementwise nonnegativity, which ensures physical interpretability, and $A^T \cdot \mathbbm{1}_p = \mathbbm{1}_S$ ensures that the fractional abundances sum to 1 for each pixel. $\mathbbm{1}_p$ and $\mathbbm{1}_S$ are column vectors of ones with size $p$ and $S$, respectively.

Supervised unmixing methods assume that endmember signatures are known a priori, allowing a sequential pipeline where abundance estimation is performed after spectral signatures are provided. However, in real-world applications, predefined endmember libraries are often insufficient due to environmental variations, sensor noise, or the presence of unknown materials. Blind unmixing methods overcome this limitation by jointly estimating both the endmember spectra and their respective abundances. This joint estimation makes blind unmixing an inherently difficult problem, often requiring additional constraints or regularization techniques to ensure meaningful solutions.

Three major classes of blind linear unmixing methods can be distinguished and are used within the proposed approach: (i) least squares-based (LS), (ii) deep learning-based (DL), and (iii) statistical-based (ST) methods. Given the inherent non-convexity of blind unmixing formulations, these methods are initialized using geometric endmember extraction techniques~\cite{hysupp}.

Each of these methods presents trade-offs in terms of accuracy, interpretability, and computational efficiency, therefore, hybrid strategies combining multiple unmixing techniques present some benefits. In the proposed approach, we integrate multiple blind linear unmixing techniques across the LS, DL, and ST categories to extract endmembers and abundance maps in an unsupervised manner. This multi-method strategy enhances robustness by leveraging the complementary strengths of different approaches. Furthermore, since these extracted abundances serve as training data for the subsequent diffusion model, using diverse unmixing techniques introduces variability into the dataset. This implicit, task-specific data augmentation improves the generalization ability of the generative model and enhances the realism of synthetic abundance maps.

\subsubsection{Least squares-based methods}

LS methods cast blind unmixing as a constrained optimization problem, minimizing the reconstruction error while enforcing spatial and geometric constraints:

\begin{align}\label{eq:ls_un}
    (\hat{A}, \hat{E}) = \arg\min_{A, E}{\frac{1}{2}||Y-EA||^2_F + \xi\phi(A)+\gamma\psi(E)}\\
    \nonumber s.t. A\geq0, 0\leq E\leq 1
\end{align}
    
\noindent where $\phi(A)$ is usually a spatial regularizer promoting spatial coherence in abundance maps and $\psi(E)$ is a geometrical regularizer ensuring physically plausible spectral signatures. Different choices of the regularizers lead to different unmixing methods~\cite{MV_bum, 1344161}. LS methods are computationally efficient and interpretable but often struggle in highly mixed or noisy environments due to their reliance on linear assumptions.

\subsubsection{Deep learning-based methods}

DL approaches typically involve autoencoder architectures to learn a compact representation of the hyperspectral data. The encoder $f_{\vartheta}$ learns to map the input image pixels into the corresponding abundance representations:

\begin{equation}\label{eq:enc_un}
    \hat{A} = f_{\vartheta}(Y)
\end{equation}
and the decoder $g_{\vartheta}$ attempts to reconstruct the original hyperspectral input image $Y$:

\begin{equation}\label{eq:dec_un}
    \hat{Y} = g_{\vartheta}(\hat{A}) = \hat{E}\hat{A}
\end{equation}

\noindent where $\vartheta$ represents the network hyperparameters and the estimated endmembers to be learned correspond to the weights of the decoder~\cite{hysupp, cnnaeu, pgmsu}. The loss function is usually based on the reconstruction result:

\begin{equation}
    \mathcal{L} = \mathcal{L}_{rec}(Y, \hat{Y})
\end{equation}

DL-based methods offer high representational power and can capture nonlinear spectral mixing effects. However, they require substantial training data, which is often limited in hyperspectral imaging, and their black-box nature can hinder interpretability.

\subsubsection{Statistical-based methods}

Statistical approaches typically employ Bayesian inference to model the distributions of endmembers and abundances~\cite{BSMA, NIPS1999_798ed7d4}. The objective is to estimate the posterior distributions of $\hat{E}$ and $\hat{A}$ within a probabilistic framework parametrized by prior distributions for the abundances $p(A)$ and the endmembers $p(E)$, and likelihood functions $p(Y|E, A)$:

\begin{equation}\label{eq:st}
    p(E,A|Y) \propto p(Y|A,E)p(A)p(E)
\end{equation}

These methods are advantageous in capturing uncertainty and incorporating domain knowledge but can be computationally demanding due to the need for iterative inference techniques.

\subsection{Diffusion model for the estimation of the abundances}

A diffusion model consists of two main components: the forward (or diffusion) process and the reverse (or denoising) process, both of which can be modeled as Markov chains~\cite{bishop}. In essence, these models allow the transformation of simple noise into complex data distributions in a way that can be both mathematically rigorous and computationally efficient.

\subsubsection{Forward process}

Let $A_0$ denote the original abundance map without any added noise, and let $q(A_0)$ represent its probability distribution. We define $A_t$ as the abundance map after $t$ successive applications of Gaussian noise. The forward process incrementally applies linear combinations of the abundance map with additive white Gaussian noise over $T$ consecutive timesteps, generating a sequence of progressively noisier abundance maps $A_0, \ldots, A_T$. This process follows a Markov chain structure, where each state $A_t$ depends solely on the previous state $A_{t-1}$ and the standard deviation of the Gaussian noise, denoted as $\beta_t$ for $t = 1, 2, \dots, T$. According to the Markov property, the probability density function (pdf) $q(A_T)$ of the forward process can be expressed as follows~\cite{bishop}:

\begin{equation}
    q(A_t|A_{t-1}) = \mathcal{N}(A_t|\sqrt{1-\beta_t}A_{t-1}, \beta_tI),
\end{equation}

\begin{equation}
    q(A_{1:T}|A_0) = \prod_{t=1}^{T}{q(A_t|A_{t-1})},
\end{equation}
where $\mathcal{N}(\cdot|\mu,\Sigma)$ denotes the Gaussian pdf with mean $\mu$ and covariance matrix $\Sigma$. The value of $A_t$ at any given time step $t$ can be expressed as a function of $A_0$ and $\beta_t$. Let us define $\alpha_t = \prod_{\tau=1}^{t}{(1-\beta_{\tau})}$~\cite{bishop}:

\begin{equation}
    A_t = \sqrt{\alpha_t}A_0 + \sqrt{1-\alpha_t}\epsilon_t
\end{equation}

\noindent where 
$\epsilon_t$ represents the additive Gaussian noise and its pdf is $\mathcal{N}(\epsilon_t|0, I)$ ($t=1,2,\ldots,T$). At this point, $q(A_t|A_0)$ can be reformulated as follows~\cite{bishop}:

\begin{equation}
    q(A_t|A_0) = \mathcal{N}(A_t|\sqrt{\alpha_t}A_0, (1-\alpha_t)I)
\end{equation}

\subsubsection{Reverse process}

The reverse process aims to reconstruct the abundance maps $A_0$ starting from the Gaussian noise $A_T$. The objective would be to compute the distribution of the reverse process $q(A_{t-1}|A_t)$, which depends on the entire data distribution and is analytically intractable~\cite{bishop}. Therefore, a DL model is employed to approximate the reverse distribution $p(A_{t-1}|A_t, w)$, where $w$ are the weights of the neural model.

\begin{equation}
    p(A_{t-1}|A_t, w) = \mathcal{N}(A_{t-1}|\mu(A_t, w, t), \beta_tI)
\end{equation}

\begin{equation}
    p(A_{0:T}|w) = p(A_T)\prod_{t=1}^{T}{p(A_{t-1}|A_t, w)},
\end{equation}

\noindent where $\mu(A_t,w,t)$ is a deep neural network governed by the vector of parameters $w$. 

The inverse process iteratively removes noise, estimated by a neural network from an initial random input. Following existing diffusion-based methods~\cite{dm, unmixingDM, 10081412, dm_gan, dm_icml}, the U-Net framework with its encoder-decoder structure has been chosen as the basic architecture enhanced by residual attention modules for feature extraction.

\subsubsection{Denoising neural network}

In the context of the reverse diffusion process, the U-Net is augmented with residual attention modules, which further enhance the model's ability to capture long-range dependencies and global context. These modules employ time embeddings to incorporate temporal information into the feature extraction process, allowing the network to adapt dynamically as the denoising progresses over the time steps.

The attention mechanism, specifically multihead self-attention, is utilized to model long-range interactions within the data. This enables the network to focus on different parts of the input at different timesteps, capturing relationships that may not be immediately apparent in localized regions of the data. Convolutional layers are used to model local dependencies, while attention layers are responsible for global context~\cite{tf_1}. 

Additionally, multiple residual connections are employed throughout the network, improving the flow of information and allowing the network to better preserve important features during the denoising process.

The overall architecture follows the typical encoder-decoder structure of the U-Net. The encoder is composed of residual blocks with self-attention layers at specific resolutions, and the spatial dimensions are progressively halved to capture hierarchical features. The bottleneck consists of two residual blocks with self-attention, enforcing the model to capture high-level abstract representations. The decoder is symmetrical to the encoder and progressively upsamples the feature maps, while incorporating the corresponding encoder features through skip connections, to preserve spatial details. The loss function is the L1 loss -- or mean absolute error (MAE) -- between the real noise and the predicted noise by the model.

The synthesis process takes as input a random noise map with the same dimensionality as the abundance maps used during training. The trained diffusion model removes the noise, gradually approaching a realistic distribution. The training and synthesis processes are summarized in algorithms~\ref{alg:alg1}-\ref{alg:alg2}.

\begin{algorithm}
\caption{Training of the diffusion model}\label{alg:alg1}
\begin{algorithmic}[1]
\State \textbf{Input:} abundance map $A_0\in\mathbb{R}^{p\times S}$ 
\State \textbf{while} not converge \textbf{do} 
    \State $(A_0,w)\sim p(A,w)$;
    \State $t\sim U(\{1,\ldots,T\})$;
    \State $\epsilon \sim \mathcal{N}(\epsilon;0,I)$;
\State Gradient descent
$ \nabla||\epsilon-\epsilon(\sqrt{\alpha_t}A_0 + \sqrt{1-\alpha_t}\epsilon,w,t)||^2$
\State \textbf{end}
\end{algorithmic}
\end{algorithm}

\begin{algorithm}
\caption{Generation of the synthetic abundance maps}\label{alg:alg2}
\begin{algorithmic}[1]
\State \textbf{Input:} random noise $A_T\in\mathbb{R}^{p\times S}$ 
\State \textbf{Output:} synthetic abundance maps $\tilde{A_0}\in\mathbb{R}^{p\times S}$ 
    \State $A_T\sim \mathcal{N}(A_T;0,I)$;
\State \textbf{for} $t=1$ to $T$ \textbf{do}
    \State $z\sim \mathcal{N}(z;0,I)$;
    \State $\tilde{A}_{t-1}=\frac{1}{\sqrt{\alpha_t}}(\tilde{A}_t-\frac{\beta_t}{\sqrt{1-\alpha_t}}\epsilon(\tilde{A}_t,w,t))+\sigma_tz$;
\State \textbf{end}
\State \textbf{return} $\tilde{A}_0$
\end{algorithmic}
\end{algorithm}
\section{Experimental Results}\label{sec:exp}
\subsection{Dataset and experimental setup}

\begin{figure*}[ht]
    \centering
    \includegraphics[width=\linewidth]{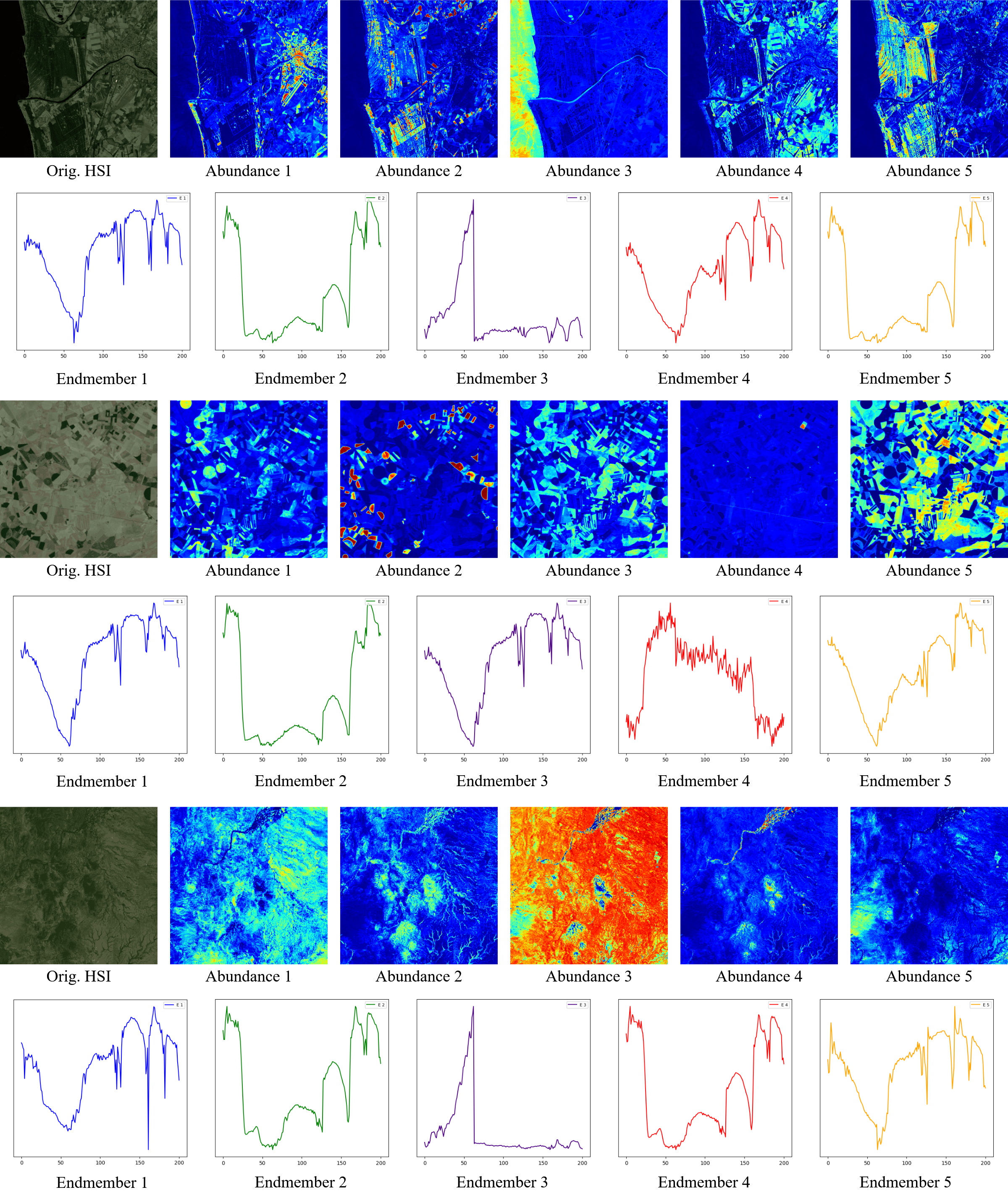}

\caption{Original PRISMA hyperspectral images (composite of bands 60, 90, 120) and unmixing results (abundances and endmembers) obtained by the convolutional autoencoder in~\cite{cnnaeu}.}
\label{fig:unmixing} 
\end{figure*}

\begin{figure*}[t]
\begin{minipage}{\textwidth}
\rotatebox{90}{\small Mixed areas}
\subfloat{\includegraphics[width=32mm]{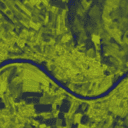}}\hfill
\subfloat{\includegraphics[width=32mm]{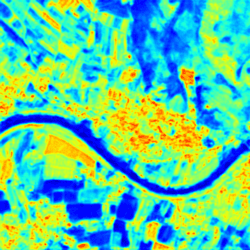}}\hfill
\subfloat{\includegraphics[width=32mm]{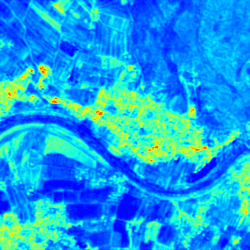}}\hfill
\subfloat{\includegraphics[width=32mm]{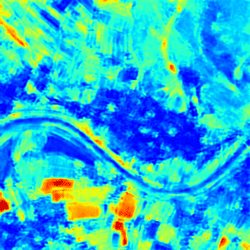}}\hfill
\subfloat{\includegraphics[width=32mm]{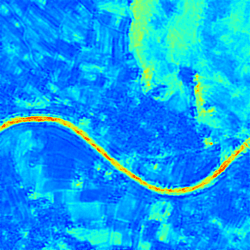}} 
\subfloat{\includegraphics[height=32mm]{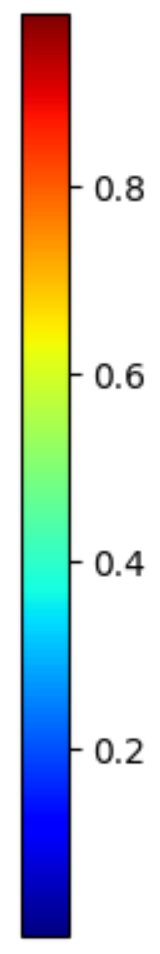}}\hfill\\
\end{minipage}
\vfill
\begin{minipage}{\textwidth}
\rotatebox{90}{\small Vegetated areas}
\subfloat{\includegraphics[width=32mm]{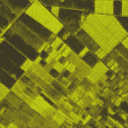}}\hfill
\subfloat{\includegraphics[width=32mm]{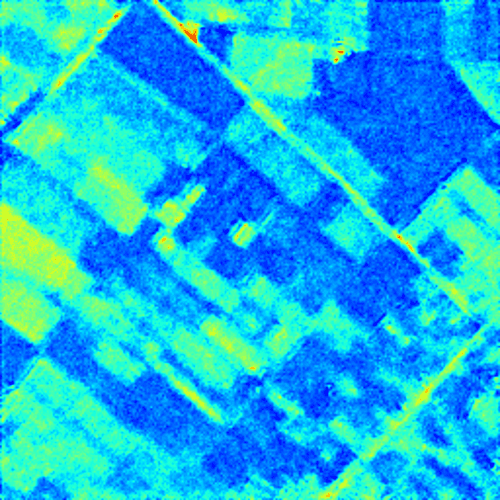}}\hfill
\subfloat{\includegraphics[width=32mm]{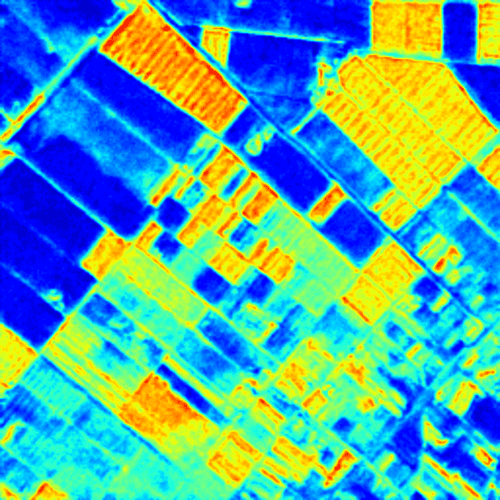}}\hfill
\subfloat{\includegraphics[width=32mm]{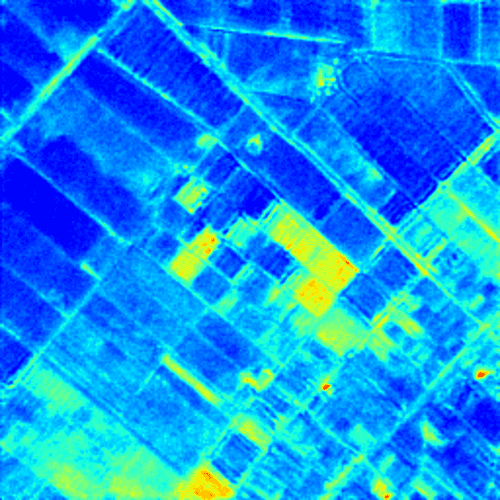}}\hfill
\subfloat{\includegraphics[width=32mm]{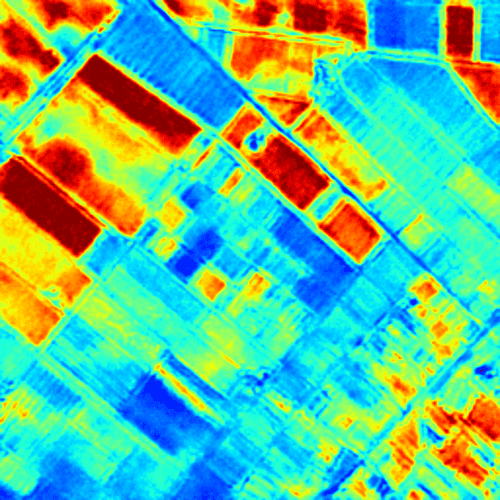}} 
\subfloat{\includegraphics[height=32mm]{Figures/colormap.PNG}}\hfill\\
\end{minipage}
\vfill
\begin{minipage}{\textwidth}
\rotatebox{90}{\small Urban areas}
\subfloat{\includegraphics[width=32mm]{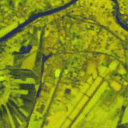}}\hfill
\subfloat{\includegraphics[width=32mm]{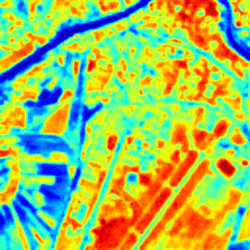}}\hfill
\subfloat{\includegraphics[width=32mm]{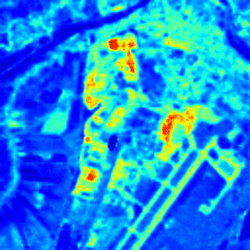}}\hfill
\subfloat{\includegraphics[width=32mm]{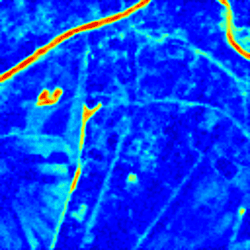}}\hfill
\subfloat{\includegraphics[width=32mm]{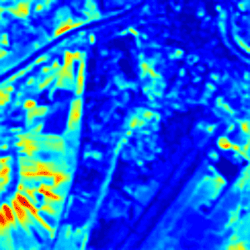}} 
\subfloat{\includegraphics[height=32mm]{Figures/colormap.PNG}}\hfill\\
\end{minipage}
\vfill
\begin{minipage}{\textwidth}
\rotatebox{90}{\small Mountainous areas}
\subfloat{\stackunder[1pt]{\includegraphics[width=32mm]{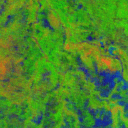}}{\small Three bands composite}}\hfill
\subfloat{\stackunder[1pt]{\includegraphics[width=32mm]{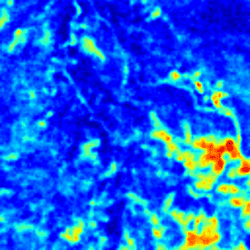}}{\small Synthetic abundance maps}}\hfill
\subfloat{\stackunder[1pt]{\includegraphics[width=32mm]{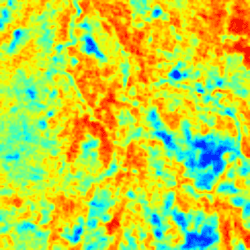}}{\small \textcolor{white}{A}}}\hfill
\subfloat{\stackunder[1pt]{\includegraphics[width=32mm]{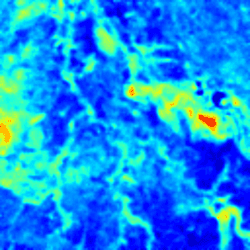}}{\small \textcolor{white}{A}}}\hfill
\subfloat{\stackunder[1pt]{\includegraphics[width=32mm]{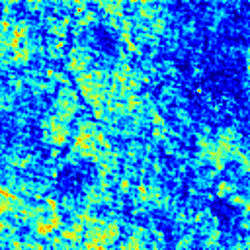}}{\small \textcolor{white}{A}}} 
\subfloat{\stackunder[1pt]{\includegraphics[height=32mm]{Figures/colormap.PNG}}{\small \textcolor{white}{A}}}\hfill\\
\end{minipage}
\caption{Examples of synthetic abundance maps obtained by the proposed method.
}
\label{fig:synth} 
\end{figure*}

The proposed method was experimentally validated using a dataset of hyperspectral images from the PRISMA mission of the Italian Space Agency. The selected acquisitions cover regions in Mexico, Spain, and Italy, forming a diverse training set that includes urban, vegetated, and mountainous landscapes. PRISMA provides both panchromatic and hyperspectral imagery; however, given our focus on hyperspectral data, only the latter was considered. The images have an average size of approximately 1200 $\times$ 1200 pixels across 234 spectral bands. After a preliminary preprocessing step to remove bands affected by atmospheric extinction, 201 spectral bands were retained. The spatial resolution of the hyperspectral data is 30 meters.  

To extract abundance maps, blind linear unmixing techniques from the three aforementioned main methodological categories were employed: least squares-based (LS)~\cite{MV_bum}, deep learning-based (DL)~\cite{cnnaeu, pgmsu}, and statistical-based (ST)~\cite{BSMA}, following the framework outlined in~\cite{hysupp}. The experiments were conducted using a variable number of endmembers, ranging from 3 to 5. Experiments performed with a higher number of endmebers yielded worse results. The resulting abundance maps, already normalized in the range $[0, 1]$ by definition, were then cropped into patches of 128 $\times$ 128 pixels, which served as training data for the diffusion model. No additional preprocessing was applied on the abundance maps extracted by the unmixing methods.

The training for the diffusion model involves an Adam optimizer, a learning rate of 1e-4, and an exponential moving average (EMA) strategy with a decay of 0.9999 for stable training. Training spans 30 million iterations. The noise schedule follows a linear beta schedule, spanning 2000 diffusion steps for both training and validation. The model was trained on an RTX4080Ti GPU.

\subsection{Results and discussion}

Fig.~\ref{fig:unmixing} illustrates the blind unmixing results obtained using the convolutional autoencoder from~\cite{cnnaeu} with five endmembers, applied to PRISMA images of both rural and urban areas. The plotted endmember spectra and corresponding abundance maps demonstrate that the method effectively differentiates the various materials present in the hyperspectral pixels, confirming that five endmembers provide a suitable representation. Experiments with a higher number of endmembers yielded suboptimal results and are omitted for brevity. Regarding the second PRISMA acquisition (second row of Fig.~\ref{fig:unmixing}), the last endmember appears only sparsely within the scene. This is evident in the corresponding abundance map (second row, last abundance map of Fig.~\ref{fig:unmixing}), where its representation is minimal. This outcome can be attributed to the nature of the material associated with the endmember—urban structures within a predominantly vegetated and cultivated landscape—occupying only a small fraction of the hyperspectral image pixels.

The extracted abundance maps were used to train the diffusion model described in Section~\ref{sec:meth}. Since the reconstruction and synthesis process relies on a U-Net-based fully convolutional network, which maintains the input size, the generated abundance map patches have a fixed size of 128 $\times$ 128 pixels. Each pixel is $p$-dimensional, where $p$ corresponds to the number of endmembers.

Fig.~\ref{fig:synth} presents examples of synthetic abundance maps corresponding to various land cover types, including mountainous, vegetated, urban, and mixed areas, consistent with the training dataset. Visual photointerpretation suggests that the generated maps exhibit a high level of realism, preserving fine spatial structures and demonstrating a logical and coherent distribution of objects across different regions. The results effectively capture the spectral and spatial diversity characteristic of real hyperspectral imagery, highlighting the effectiveness of the proposed approach. Additionally, the generated maps maintain a smooth and spatially coherent transition between different materials, avoiding abrupt artifacts that could indicate synthetic inconsistencies.

For mountainous scenes, a slight presence of Gaussian noise is observed in one of the abundance maps. However, this residual noise remains localized and does not significantly affect the overall interpretability of the results. Notably, such artifacts are absent in the abundance maps generated for other land cover types, indicating that the diffusion model has successfully generalized across different environmental conditions. This suggests that, despite minor imperfections, the model has learned to generate abundance maps that align well with the statistical properties of real data.

Overall, the generated synthetic maps provide a meaningful approximation of real-world abundance distributions, demonstrating the capability of the proposed approach to model significant realizations of the underlying random field of abundances.

\section{Conclusions}

This paper introduces a methodology for synthesizing abundance maps by integrating a plethora of blind linear unmixing algorithms with deep diffusion models. The proposed approach leverages blind linear unmixing to extract abundance maps from real hyperspectral acquisitions for training purposes, while a deep diffusion model is employed to efficiently learn and generate realistic samples from the underlying abundance distribution.  

The experimental analysis was conducted using a training set of abundance maps extracted from PRISMA hyperspectral images acquired over Mexico, Spain, and Italy. Visual inspection of the results suggests that the proposed method effectively generates realistic abundance maps corresponding to diverse land cover types, including mountainous, vegetated, urban, and mixed environments. These findings highlight the potential of diffusion models as powerful generative frameworks and demonstrate the effectiveness of the proposed approach in utilizing these models for abundance map synthesis within hyperspectral image modeling. This has potential applications in sensor simulation, mission planning, and training data augmentation for hyperspectral imaging research.  

Future work will extend the method to full hyperspectral image synthesis by incorporating the learned abundance model. The generated hyperspectral images will be validated with quantitative metrics such as spectral angle distance (SAD), root mean square error (RMSE), peak signal to noise ratio (PSNR), and structural similarity index (SSIM). Additionally, an interesting avenue for further exploration is the integration of super-resolution techniques to enable the generation of abundance maps with higher spatial resolution. In this regard, super-resolution diffusion models~\cite{9887996} will be considered to enhance the spatial fidelity of the synthetic hyperspectral data.

\small\section{Acknowledgment}
This work was partially supported by the Italian Space
Agency (ASI) within the Framework of the Project under Grant SIM4PRISMA2G-ASI no. 2023-2-HB.0. PRISMA Products, ©ASI, delivered under a license to use by ASI.

{
    \small
    \bibliographystyle{ieeetr}
    \bibliography{main}
}


\end{document}